\documentclass[10pt,twocolumn,letterpaper]{article}

\usepackage{cvpr}
\usepackage{times}
\usepackage{epsfig}
\usepackage{graphicx}
\usepackage{amsmath}
\usepackage{amssymb}
\usepackage{color}
\usepackage{booktabs}
\usepackage{multirow}

\usepackage[pagebackref=true,breaklinks=true,colorlinks,bookmarks=false]{hyperref}

\cvprfinalcopy 

\def\cvprPaperID{237} 

\DeclareMathOperator*{\argmax}{argmax}
\DeclareMathOperator*{\argmin}{argmin}

\begin{document}

\title{Real-Time 3D Model Tracking in Color and Depth on a Single CPU Core}

\author{
Wadim Kehl $^{1,2}$ \hspace{0.5cm} Federico Tombari $^{2}$ \hspace{0.5cm} Slobodan Ilic $^{2,3}$ \hspace{0.5cm} Nassir Navab $^{2}$ \\
$^{1}$ Toyota Research Institute, Los Altos, USA  \hspace{0.5cm} $^{2}$ Technical University of Munich, Germany \\
$^{3}$ Siemens Corporate Research, Munich, Germany \\
{\tt\small wadim.kehl@tri.global}
}

\maketitle

\begin{abstract}
	
	We present a novel method to track 3D models in color and depth data. To this end, we introduce approximations that accelerate the state-of-the-art in region-based tracking by an order of magnitude while retaining similar accuracy. Furthermore, we show how the method can be made more robust in the presence of depth data and consequently formulate a new joint contour and ICP tracking energy. We present better results than the state-of-the-art while being much faster then most other methods and achieving all of the above on a single CPU core. 

\end{abstract}

\section{Introduction}

Tracking objects in image sequences is a relevant problem in computer vision with significant applications in several fields, such as robotics, augmented reality, medical navigation and surveillance. For most of these applications, object tracking has to be carried out in 3D, $\ie$ an algorithm has to retrieve the full 6D pose of each model in every frame. This is quite challenging since objects can be ambiguous in their pose and can undergo occlusions as well as appearance changes. Furthermore, trackers must also be fast enough in order to cover larger inter-frame motions.

In the case of 3D object tracking from color images, the related work can be roughly divided into sparse methods that try to establish and track local correspondences between frames \cite{Vacchetti2004, Park2008}, and region-based methods that exploit more holistic information about the object such as shape, contour or color \cite{Prisacariu2012,Dambreville2010}, although mixtures of both do exist \cite{Schmaltz2007,Brox2010, Schmaltz2012}. While both directions have their respective advantages and disadvantages, the latter performs better for texture-less objects, which is our focus here. One popular methodology for texture-less object tracking relies on the idea of aligning the projected object contours to a segmentation in each frame. While initially shown for arbitrary shapes \cite{Cremers2007,Bibby2008}, more recent works put emphasis on tracking 3D colored models \cite{Dambreville2010,Prisacariu2012,Zhao2014,Tjaden2016}. 

\begin{figure}
	\centering
	\includegraphics[width=8.2cm]{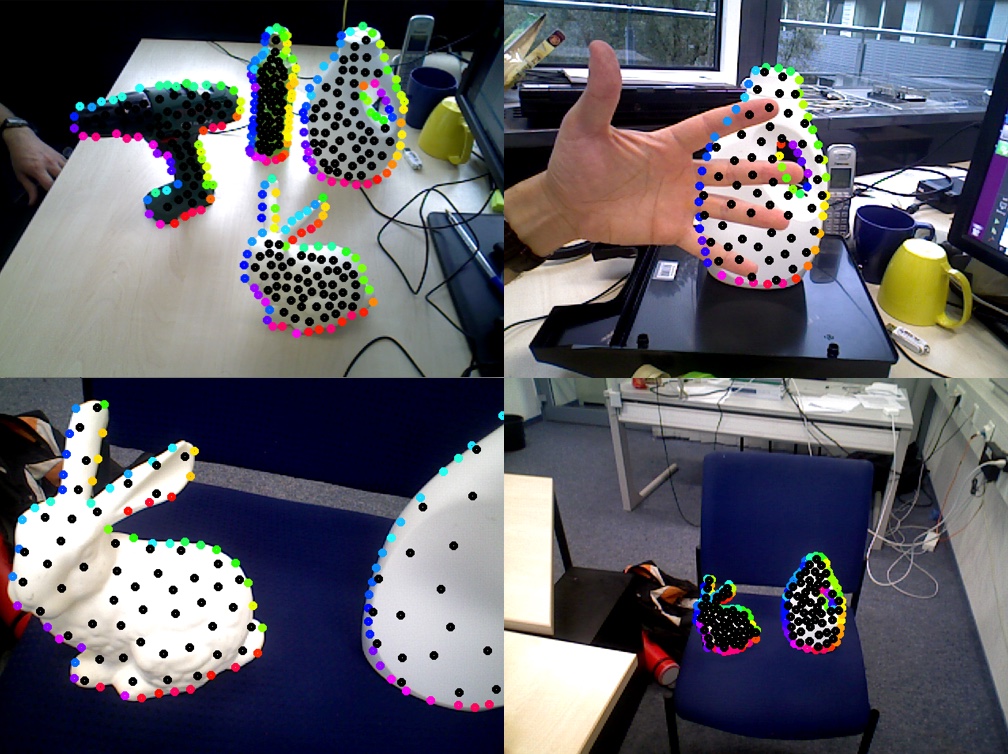} 
	\caption{We can perform reliable tracking for multiple objects while being robust towards partial occlusions as well drastic changes in scale. To this end, we employ contour cues and interior object information to drive the 6D pose alignment jointly. All of the above is achieved on a single CPU core in real-time. }
	\label{fig:intro}
\end{figure}

With the advent of commodity RGB-D sensors, these methods have then been further extended to depth images for simultaneous tracking and reconstruction \cite{Ren2013,Ren2014}. Indeed, exploiting RGB-D is beneficial since image-based contour information and depth maps are complimentary cues, one  being focused on object borders, the other on object-internal regions. This has been exploited for 3D object detection and tracking \cite{Park2011,Hinterstoisser2012a,Kehl2015}, as well as to improve camera tracking in planar indoor environments \cite{Zhou2015}.

From a computational perspective, several state-of-the-art trackers leverage the GPU for real-time performance \cite{Ren2013,Ren2014,Tjaden2016}. Nevertheless, there is a strong interest towards decreasing the computational burden and generally avoiding GPU usage, motivated by the fact that many relevant applications require trackers to be light-weight \cite{Holzer2012,Prisacariu2015}.

Taking this all into consideration, we propose a framework that allows accurate tracking of multiple 3D models in color and depth. Unlike related works~\cite{Prisacariu2012,Tjaden2016} our method is lightweight, both in computation (requiring only one CPU core) and in memory footprint. To achieve this, we propose to pre-render a given target 3D model from various viewpoints and extract occluding contour and interior information in an offline step. This avoids time consuming online renderings and consequently results in a fast tracking approach. Furthermore, we do not compute the terms of our objective function densely but introduce sparse approximations which gives a tremendous performance boost, allowing real-time tracking of multiple instances. While the proposed contour-based tracking works well in RGB images, in the case of available depth information, we propose two additions: firstly, we make the color-based segmentation more robust by incorporating cloud information and secondly, we define a new tracking framework where a novel plane-to-point error on cloud data and a contour error are simultaneously steering the pose alignment.  

\begin{itemize}
\item As a foundation of our work, we propose to pre-render the model view space and extract contour and interior information in an offline step to avoid online rendering, making our method a pure CPU-based approach.
\item We evaluate all terms sparsely instead of densely which gives a tremendous performance boost.
\item Given RGB-D data, we show how to improve contour-based tracking by incorporating cloud information into the color contour estimation. Additionally, we present a new joint tracking that incorporates a novel plane-to-point error and a contour error, \ie color and depth points are simultaneously steering the pose alignment.
\end{itemize}

Therefore, our method can deal with challenges typically encountered in tracking as depicted in the Figure \ref{fig:intro}. In the results section, we evaluate our approach both quantitatively and qualitatively and compare it to the related approaches reporting better accuracy at higher speeds.

\section{Related work}

We confine ourselves to the field of 3D model tracking in color and depth. Earlier works in this field employ either 2D-3D correspondences \cite{Rosenhahn2006,Schmaltz2007} or 3D edges \cite{Drummond2002,Tateno2009, Seo2014} and fit the model in an ICP fashion, i.e. without explicitly computing a contour. While successive methods along this direction managed to obtain improved performance \cite{Brox2010,Schmaltz2012}, another set of works solely focused on tracking densely the contour by evolving a level-set function \cite{Bibby2008,Dambreville2010}. In particular, Bibby \etal \cite{Bibby2008} aligned the current evolving 2D contour to a color segmentation, and demonstrated improved robustness when computing a posterior distribution in color space. 

Based on this work, the first real-time contour tracker for 3D models was presented by Prisacariu \etal \cite{Prisacariu2012}, where the contour is determined by projecting the 3D model with its associated 6D pose onto the frame. Then, the alignment error between segmentation and projection drives the update of the pose parameters via gradient descent. In a follow-up work \cite{Prisacariu2015}, the authors extend their method to simultaneously track and reconstruct a 3D object on a mobile phone in real-time. They circumvent GPU rendering by hierarchically ray-casting a volumetric representation and speed up pose optimization by exploiting the phone's inertial sensor data. 
Tjaden \etal \cite{Tjaden2016} build on the original framework and extend it with a new optimization scheme that employs Gauss-Newton together with a twist representation. Additionally, they handle occlusions in a multi-object tracking scenario, making the whole approach more robust in practice. 
The typical problem of these methods is their fragile segmentation based on color histograms, which can fail easily without using an adaptive appearance model, or when tracking in scenes where the background colors match the objects' colors. Based on this, \cite{Zhao2014} explores a boundary term to strengthen contours, whereas \cite{Hexner2016} improves the segmentation with local appearance models. 

When it comes to temporal tracking from depth data, there are mostly only works based on energy minimization of sparse and dense data \cite{Newcombe2011,Choi2013,Ren2013,Ren2014,Zhou2015, Slavcheva2016}, or based on learning, such as the work from Tan \etal \cite{Tan2014,Tan2015} and Krull \etal \cite{Krull2014}. Among these, the closest to us are the works from Ren \etal \cite{Ren2013,Ren2014}, which track and simultaneously reconstruct a 3D level-set embedding from depth data, following a color-based segmentation. We move orthogonally by looking at depth as an additional modality towards reliable segmentations and to improve tracking via a novel ICP term in a joint formulation. 

\section{Methodology}
\begin{figure*}[t]
	\centering
	\includegraphics[width=17cm]{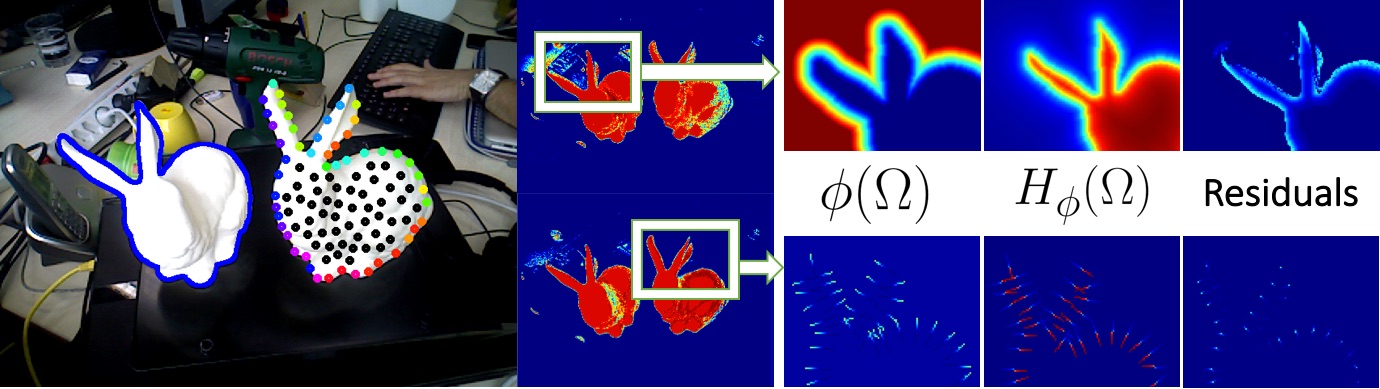} 
	\caption{Tracking two Stanford bunnies side by side in color data. While the left is tracked densely, the right is tracked with our approximation via a sparse set of 50 contour sample points. Starting from a computed posterior map $P_f$ for each object, we depict some involved energy terms. The color on each sparse contour point represents its 2D orientation whereas the black dots are the sampled interior points. }
	\label{fig:teaser}
\end{figure*}

We will first introduce the notion of contour tracking in RGB-D images. There we formalize a novel foreground posterior probability composed of both color and cloud data. This is followed by the complete energy formulation over joint contour and cloud alignment. Finally, we then explain our further contributions to boost runtime performance via our proposed approximation schemes.

\subsection{Tracking via implicit contour embeddings}

In the spirit of \cite{Prisacariu2012,Tjaden2016}, we want to track a (meshed) 3D model in camera space such that its projected silhouette aligns perfectly with a 2D segmentation in the image $I: \Omega \rightarrow \mathbb{R}^3$ with $\Omega \subset \mathbb{R}^2$ being the image domain. 
Given a silhouette (\ie foreground mask) $\Omega_f \subseteq \Omega$, we can infer a contour $C$ to compute a signed distance field (SDF) $\phi$ s.t.
\begin{equation}
\phi(x):= 
\begin{cases}
 d(x,C), & x \in \Omega_b \\
-d(x,C), & x \in \Omega_f
\end{cases}
, \text{ } d(x,C) := \min_{y \in C} ||x-y||
\label{eq:sdf}
\end{equation}
where a pixel tells the signed distance to the closest contour point and $\Omega_b := \Omega \setminus \Omega_f$ is the set of background pixels.

We follow the PWP3D tracker energy formulation \cite{Prisacariu2012} in which the pixel-wise (posterior) probability of a contour, embedded as $\phi$, given color image $I$, is defined as
\begin{equation}
P(\phi|I) := \prod_{x \in \Omega} \bigg(  H_\phi(x) P_f(I(x)) +  (1-H_\phi(x) P_b(I(x)) ) \bigg).
\label{eq:contour_prob}
\end{equation}
The terms $P_f, P_b$ are modeling posterior distributions for foreground and background membership based on color, in practice computed from normalized RGB histograms, whereas $H_\phi$ represents a smoothed version of the Heaviside step function defined on $\phi$. To get an impression of the involved terms, we refer to Figure \ref{fig:teaser}.

In practice, this posterior works well in cases where the foreground and background are different and starts failing when the color of the parts of the background get close to the color of the target object. To circumvent this problem, we propose to use depth information coming from the RGB-D sensor at the sparse sample points on the foreground of the object as supplementary information.

Let us define a depth map $D: \Omega \rightarrow \mathbb{R}^+$ and its cloud map $\Pi_D^{-1}: \Omega \rightarrow \mathbb{R}^3$. Furthermore, we conduct a fast depth map inpainting such that we remove all unknown values in both $D$ and $\Pi_D^{-1}$. Our goal is now two-fold: we want to make the posterior image $P_f$ more robust by including cloud information into the probability estimation, and we want to extend the tracking energy to the new data.

\subsection{Pixel-wise color/cloud posterior}
In practice, color histograms are very error prone and fail quickly for textured/glossy objects and colorful backgrounds, even with adaptive histogram during tracking. We therefore propose a new robust pixel-wise posterior $P_f$ to be used for contour alignment in Eq. \ref{eq:contour_prob} when additional depth data is provided. The notion we bring forward is that color posteriors alone are misleading and should be reweighted with their spatial proximity to the model. 
\begin{figure}
	\centering
	\includegraphics[width=8.5cm]{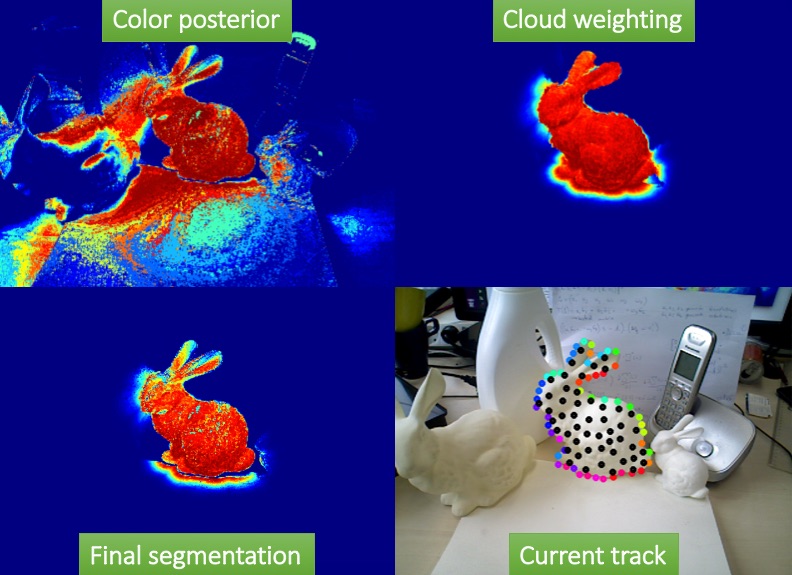} 
	\caption{Segmentation computation. Since the background is similar in color, only the additional cloud-based weighting can give us a reliable segmentation to track against.}
	\label{fig:cloud_prior}
\end{figure}
Given a model with pose $M=[R,t]$, we infer silhouette region $\Omega_f$ and background $\Omega_b$ and now define their probabilities not only based on a given pixel color $x$ but also an associated cloud point $C$. We start from estimating the probability of a pose and its silhouette, provided color and cloud data, and define $G \in \{FG,BG\}$ as a binary foreground/background variable. For tractability, we assume that a pose and its silhouette are independent, given $x$ and $C$:
\begin{equation*}
P_f:=P(G=FG, M |x,C) = P(FG|x,C) \cdot P(M|x,C).
\end{equation*}
Assuming that all pixels are independent, that there is no correlation between the color of a pixel and its cloud point, and that $P(M)$ is uniform, we reach\footnote{We refer the reader to the supplement for the full derivation.}:
\begin{eqnarray}
P(FG|x,C) & := & \frac{P(x|FG) \cdot P(C|FG) }{ \Sigma_{G \in \{FG,BG\} } P(x|G) \cdot P(C|G)} \\
P(M|x,C) & \propto & P(x|M) \cdot P(C|M).
\end{eqnarray}
While $P(x|G)$ are usually computed from color histograms, it is not directly clear of how to compute $P(C|G)$ since it assumes inference for 3D data from an image mask while the term $P(x|M)$ is infeasible in general. We thus drop both terms (\ie set both to uniform) and finally define

\begin{equation}
P_f := P(C|M) \cdot \frac{P(x|FG) }{ \Sigma_{G \in \{FG,BG\} } P(x|G)}.
\end{equation}
The weighting term $P(C|M)$, which gives the likelihood of a cloud point to be on the model, can be computed in multiple ways. Instead of simply taking the distance to the model centroid, we want a more precise measure that gives back the distance to the \textbf{closest} model point. Since even logarithmic nearest-neighbor lookups would be costly here, we use an idea first presented in \cite{Fitzgibbon2001}. One can pre-compute a distance transform in a volume around the model to facilitate a constant nearest-neighbor lookup function, $N(C) := \argmin_{X \in Model} ||X-C||$, and we exploit this approach by bringing each scene cloud point $C$ into the local object frame and efficiently calculate a pixel-wise weighting on the image plane with a Gaussian kernel: 
\begin{equation}
P(c|M) := \exp ( - \frac{||\bar{C} - N(\bar{C})||}{ \sigma^2} ) , \bar{C} := R^\intercal \cdot C - R^\intercal t.
\end{equation}
Here, $\sigma=2.5cm $ steers how much deviation we allow a point to have from a perfect alignment since we want to deal with pose inaccuracies as well as depth noise.

We can see the color posterior at work plus combination of the cloud-based weighting term in Figure \ref{fig:cloud_prior}. While the former gives a segmentation based on appearance alone, the latter takes complementary spatial distances into account, rendering contour-based tracking more robust.

\subsection{Joint contour and cloud tracking}
We introduce the notion of a combined tracking approach where 2D contour points and 3D cloud points are jointly driving the pose update. In essence, we seek a weighted energy of the form
\begin{equation}
E_{Joint} = E_{C} + \lambda E_{ICP}  .
\label{eq:joint_energy}
\end{equation}
where $\lambda$ is balances both partial energies since they can deviate in the number of samples as well as numerical scale. We want to mention the work \cite{Park2011} which formulate a similar optimization problem.

\subsubsection{Contour energy}
Assuming pixel-wise independence and taking the negative logarithm of Eq. \ref{eq:contour_prob}, we get a contour energy functional
 \begin{equation}
 E_{C} := - \sum_{x \in \Omega} \log \bigg(  H_\phi(x)  P_f(I(x)) +  (1-H_\phi(x)  P_b(I(x)) ) \bigg).
 \label{eq:energy}
 \end{equation}
In order to optimize the energy in respect to a change in model pose, we employ a Gauss-Newton scheme over twist coordinates, similarly to Tjaden et al. \cite{Tjaden2016}. We define a twist vector $\xi = [t_x,t_y,t_z, \omega_x, \omega_y,\omega_z]^\intercal \in \mathbb{R}^6$ that provides a minimal representation for our sought transformation and its Lie algebra twist $\hat{\xi} \in se(3)$ as well as its exponential mapping to the Lie group $\Xi \in SE(3)$:
\begin{equation}
\hat{\xi} := 
\begin{pmatrix}
0 & -\omega_z  & \omega_y  & t_x \\
\omega_z & 0 & -\omega_x  & t_y \\
-\omega_y  & \omega_x  & 0 & t_z \\
0 & 0 & 0 & 0
\end{pmatrix}
,
\Xi := \exp(\hat{\xi} ) = 
\begin{pmatrix}
R & t \\
0 & 1
\end{pmatrix}.
\end{equation}
We abuse notation s.t. $\Xi(X)$ expresses the transformation of $\Xi \in \mathbb{R}^{4 \times 4}$ applied to a 3D point $X$. Assuming only infinitesimal change in transformation we derive the energy\footnote{For brevity, we moved the full derivation into the supplement.} in respect to a point $X$ undergoing a screw motion $\xi$ as
\begin{equation}
\frac{ \partial E_C }{\partial \xi   } = 
\frac{  (P_f - P_b) }{H_\phi (P_f - P_b) + P_b   } 
\frac{ \partial H_\phi }{\partial \phi   }   
\frac{ \partial \phi }{\partial x   }  
\frac{ \partial \pi(X) }{\partial X   } 
\frac{ \partial \Xi(X) }{\partial \xi   } .
\label{eq:derivative}
\end{equation}
A visualization of some terms can be seen in Figure \ref{fig:teaser}. While $\frac{ \partial \Xi(X) }{\partial \xi} \in \mathbb{R}^{3 \times 6} $ and $\frac{ \partial \pi(X) }{\partial X} \in \mathbb{R}^{2 \times 3} $ can be written in analytical form, $\frac{ \partial H_\phi }{\partial \phi}$ resolves essentially to a smoothed Dirac delta whereas $\frac{ \partial \phi }{\partial x} \in \mathbb{R}^{1 \times 2} $ can be implemented via simple central differences. In total, we obtain one Jacobian $J_x \in \mathbb{R}^{1 \times 6}$ per pixel and solve a least-squares problem
\begin{equation}
\nabla \xi =  (\sum_x J_x^\intercal J_x) ^{-1}  \sum_x J_x
\label{eq:normal}
\end{equation}
via Cholesky decomposition. Given the model pose $M^t \in \mathbb{R}^{4 \times 4}$ at time $t$, we update via the exponential mapping
\begin{equation}
M^{t+1} = \exp( \hat{ \nabla \xi }) \cdot M^t.
\end{equation}

\subsubsection{ICP Energy} 
In terms of ICP, a point-to-plane error has been shown to provide better and faster convergence then a point-to-point metric. It assumes alignment of source points $s_i$ (here from a model view) to points $d_i$ and normals $n_i$ at the destination (here the scene). Normals in camera space can be approximated from depth images \cite{Hinterstoisser2012a} but are usually noisy and require time. We thus propose a novel plane-to-point error where the normals are coming from the source point set and have been computed beforehand for each viewpoint. This ensures a fast runtime and perfect data alignment since tangent planes coincide at the optimum.

Given the current pose $[R,t]$ and closest viewpoint with local interior points $s_i$, we transform to $\bar{s}_i = R \cdot s_i + t$ and project each to get the corresponding scene point $d_i := \Pi_D^{-1}(\pi(\bar{s}_i) )$. Since we also have a local $n_i$ that we bring into the scene, $\bar{n}_i = R \cdot n_i$, we want to retrieve $\Xi$ minimizing
\begin{equation}
E_{ICP} := \argmin_{\Xi} \sum_i \bigg ( (\Xi(\bar{s}_i)  - d_i) \cdot  \Xi_{SO}(\bar{n}_i)  \bigg ) ^2.
\label{eq:icp}
\end{equation}
The difference to the established point-to-plane error is solving for an additional rotation of the source normal $\bar{n}_i$. Note that only the rotational part of $\Xi$ acts on $\bar{n}_i$ and we thus omit the translational generators of the Lie algebra. Deriving in respect to $\xi$\footnote{The derivation can be found in the supplementary material.}, we get a Jacobian $J_i \in \mathbb{R}^{1 \times 6}$ and a residual $r_i$ for each correspondence
\begin{equation}
J_i  := - \bigg[ \hspace{0.25cm} \bar{n}_i^\intercal  \hspace{0.5cm} \bigg( (\bar{s}_i \times \bar{n}_i) + \bar{n}_i \times (\bar{s}_i-d_i)  \bigg)^\intercal \bigg] , 
\end{equation}
\begin{equation}
r_i := (\bar{s}_i-d_i) \cdot \bar{n}_i
\end{equation}
and construct a normal system to get a twist of the form
\begin{equation}
\nabla \xi =  \bigg( \sum_i J_i^\intercal J_i \bigg ) ^{-1}  \sum_i J_i \cdot r_i
\label{eq:normal_icp}
\end{equation}
Altogether, we can now plug together Eqs. \ref{eq:energy} and \ref{eq:icp} to formulate the desired energy from Eq. \ref{eq:joint_energy} as a joint contour and plane-to-point alignment. Following up, we build a normal system that contains both the ray-wise contour Jacobians $J_x$ from 2D image data and correspondence-wise ICP Jacobians $J_i$ from 3D cloud data:
\begin{equation}
\nabla \xi = 
\bigg( \sum_x J_x^\intercal J_x + \sum_i \lambda  J_i^\intercal J_i \bigg)^{-1} \bigg( \sum_x J_x + \sum_i \sqrt{\lambda} J_i \cdot r_i \bigg).
\end{equation}
Solving the above system yields a twist with which the current pose is updated. The advantage of such a formulation is that we employ entities from different optimization problems into a common framework: while the color pixels minimize a projective error, the cloud points do so with a geometrical error. These complimentary cues can therefore compensate for each other if a segmentation is partially wrong or if some depth values are noisy.

\subsection{Approximating for real-time tracking}

\begin{figure}
	\centering
	\includegraphics[width=8.5cm]{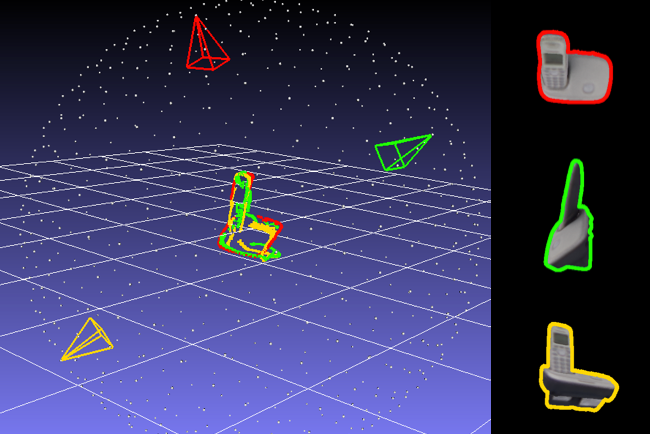} 
	\caption{Object-local 3D contour points visualized for three viewpoints on the unit sphere. Each view captures a different contour which is used during tracking to circumvent costly renderings.}
	\label{fig:contour_space}
\end{figure}

\begin{figure}
	\centering
	\includegraphics[width=8.5cm]{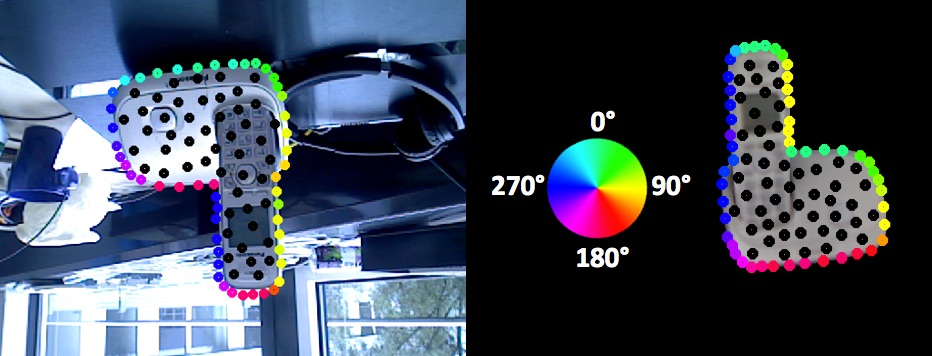} 
	\caption{Current tracking and closest pre-rendered viewpoint augmented with contour and interior sampling points. The hue represents the normal orientation for each contour point. Note how we rotate the orientation of each contour point by our approximation of the inplane rotation such that the SDF computation is proper.}
	\label{fig:closest_view}
\end{figure}

Computing the SDF from Eq. \ref{eq:sdf} has already three costly steps. We need a silhouette rendering $\Omega_f$ of the current model pose, an extraction of the contour $C$ and lastly, a subsequent distance transform embedding $\phi$. While \cite{Tjaden2016} perform GPU rendering and couple computation of the SDF and its gradient in the same pass to be faster, \cite{Prisacariu2015} perform hierarchical ray-tracing on the CPU and extract the contour via Scharr operators. We make two key observations:
\begin{enumerate}
\item Only the actual contour points are required
\item Neighboring points provide superfluous information because of similar curvature 
\end{enumerate}
We thus propose a cheap yet very effective approximation of the model render space that avoids both online rendering and contour extraction. In an offline stage, we equidistantly sample viewpoints $V_i$ on a unit sphere around the object model, render from each and extract the 3D contour points to store view-dependent \textbf{sparse} 3D sampling sets in \textbf{local object space} (see Fig. \ref{fig:contour_space}). Since we will utilize these points in 3D space, we neither need to sample in scale nor for different inplane rotations. Finally, we also store for each contour point its 2D gradient orientation and sample a set of interior surface points with their normals (see Fig. \ref{fig:closest_view}).

In a naive approach, all involved terms from Eq. \ref{eq:energy} would be computed densely, \ie $\forall x \in \Omega$, which is prohibitively costly for real-time scenarios. The related work evaluates the energy only in a narrow band around the contour since the residuals decay quickly when leaving the interface. We therefore propose to compute Eq. \ref{eq:derivative} in a narrow band along a sparse set of selected contour points where we compute $\phi$ along rays. Each projected contour point shoots a positive and negative ray perpendicularly to the contour, i.e along its normal. Building on that, we introduce the idea of ray integration for 3D contour points such that we do not create pixel-wise but \textbf{ray-wise} Jacobians which leads to a smaller reduction step and a better conditioning of the normal system in Eq. \ref{eq:normal} than \cite{Prisacariu2015} and their approach.

To formalize, we have a model pose $[R,t]$ during tracking and avoid rendering by computing the camera position in object space $O := -R^\intercal t$. We normalize to unit length and find the closest viewpoint $V^*$ quickly via dot products:
\begin{equation}
V^* := \argmax_{V_i}  \langle ~ V_i , ~ O/||O|| ~ \rangle.
\end{equation}
Each local 3D sample point of the contour $X_i$ from $V^*$ is then transformed and projected to a 2D contour sample point $x_i = \pi(R X_i + t)$ which is then used to shoot rays into the object interior and into the opposite direction. 

To get the orientation of each ray, we cannot rely anymore on the value during pre-rendering since the current model pose might have an inplane rotation not accounted for. Given a contour point with 2D rotation angle $\theta$ during pre-rendering, we could embed it into 3D space via $v = (\cos \theta, \sin \theta, 0)$ and later multiply it with the current model rotation $R$. Although this works in practice, the projection of $R\cdot v$ onto the image plane can be off at times. We thus propose a new approximation of the inplane rotation where we seek to decompose $R = R_{inplane} \cdot R_{canonical}$ s.t. one part describes a general rotation around the object center in a canonical frame and the other a rotation around the view direction of the camera (\ie inplane) . Although ill-posed in general, we exploit our knowledge about the closest viewpoint by assuming  $R_{canonical} \approx R_{V^*}$  and propose to approximate a rotation $\tilde{R}$ on the xy-plane via
\begin{equation}
\tilde{R} := R \cdot R_{V^*}^\intercal.
\end{equation}
We then extract the angle $\theta = acos(\tilde{R}_{1,1})$ via the first element. With larger viewpoint deviation  $||V^* - \frac{O}{||O||}||$, this approximation worsens but our sphere sampling is dense enough to alleviate this in practice. We re-orient each contour gradient $\tilde{g_i} := (g_i + \theta) \mod 2 \pi$ and shoot rays to compute the residuals and $\frac{ \partial H_\phi }{\partial \phi}$ from Eq. \ref{eq:energy} (see Fig. \ref{fig:closest_view} to compare the orientations and the bottom row in Figure \ref{fig:teaser} for the SDF rays).

The final missing building block is the derivative of the SDF $\frac{ \partial \phi }{\partial x }$ which cannot be computed numerically since we are missing dense information. We thus compute it geometrically, similar to \cite{Prisacariu2015}. Whereas their computation is exact when assuming local planarity by projections onto the principal ray, our approach is faster while incurring a small error which is negligible in practice. Given a ray $r = (r_x,r_y)$ from contour point $p = (p_x, p_y)$ we compute the horizontal derivative at $\phi(p_x+r_x,p_y+r_y)$ as central difference
\begin{equation}
\frac{||(p_x+r_x + 1, p_y+r_y)|| - ||(p_x+r_x -1, p_y+r_y)||}{2}.
\end{equation}
The vertical derivative is computed analogously. Like the related work, we perform all computations on three pyramid levels in a coarse-to-fine manner and shoot the rays in a band of 8 steps on each level. Since we shoot two rays per contour point, our resulting normal system holds two ray Jacobians per point.

\subsection{Implementation details}
Our method runs in C++ on a single core of an i7-5820K@3.3GHz. In total, we render a model from equidistant 642 views, amounting to around 8 degrees in angular difference between two viewpoints. To compute the histograms we avoid rendering and instead fetch the colors at the projected interior points for the foreground histogram. For the background histogram, we compute the rectangular 2D projection of the model's 3D bounding box and take the pixels outside of it. We employ 1D lookup tables for both $H_{\phi}$ and its derivative to speed up computation. Lastly, if we find a projected transformed point $\bar{p}$ to be occluded, \ie $D(\pi(\bar{p})) + 5cm < \bar{p}_z $, we discard this point for all computations.

\section{Evaluation}

To provide quantitative numbers and to self-evaluate our method on noise-free data, we run the first set of experiments on the synthetic RGB-D dataset of Choi and Christensen \cite{Choi2013}. It provides four sequences of 1000 frames where each covers an object around a given trajectory. Later, we run convergence experiments on the LineMOD dataset \cite{Hinterstoisser2012a} and evaluate against Tan \etal on two of their sequences.


\subsection{Balancing the tracking energy with $\lambda$}
\begin{figure}
	\centering
	\includegraphics[width=4cm]{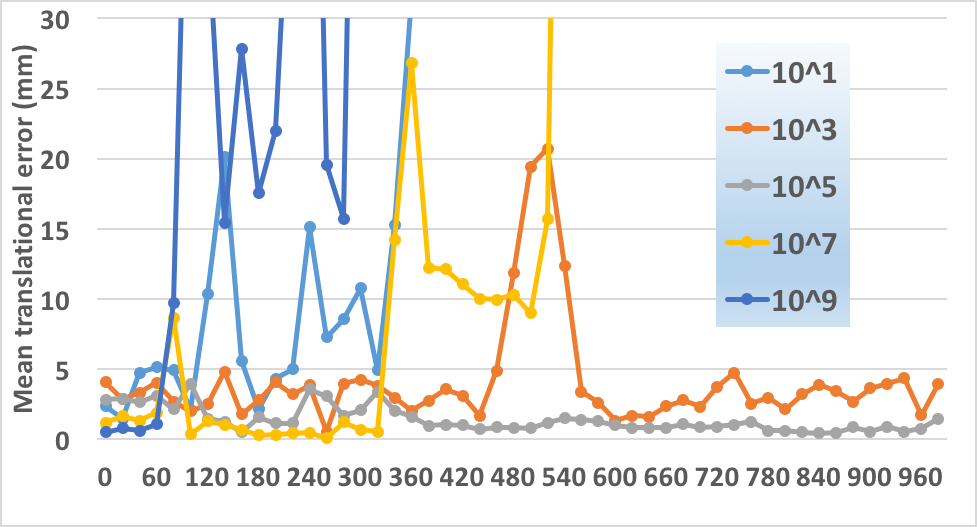} 
	\includegraphics[width=4cm]{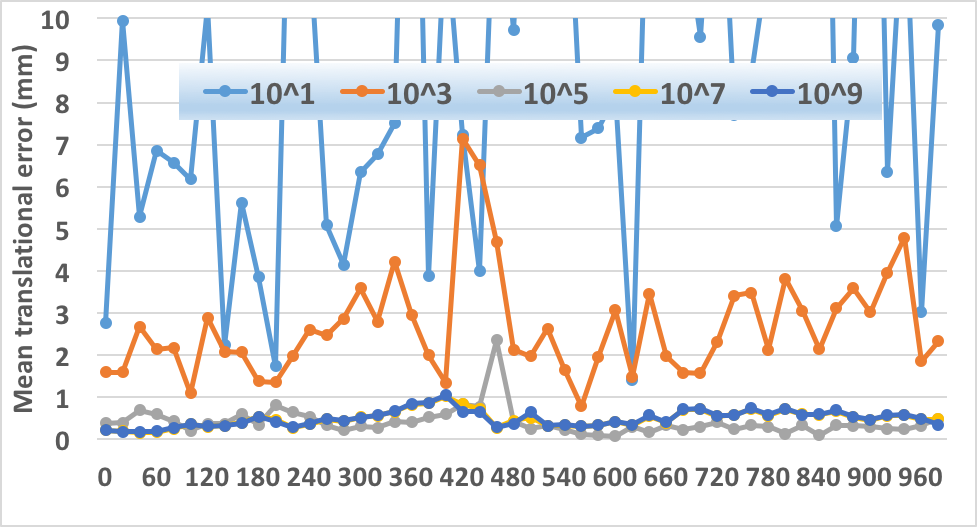} \\
		\includegraphics[width=2.5cm]{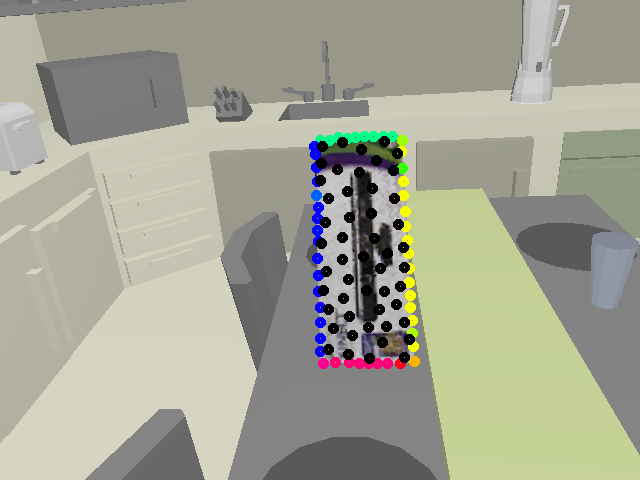} 
		\includegraphics[width=2.5cm]{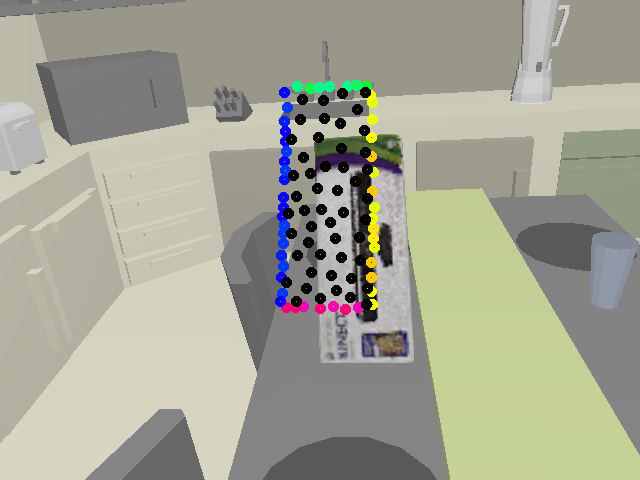} 
		\includegraphics[width=2.5cm]{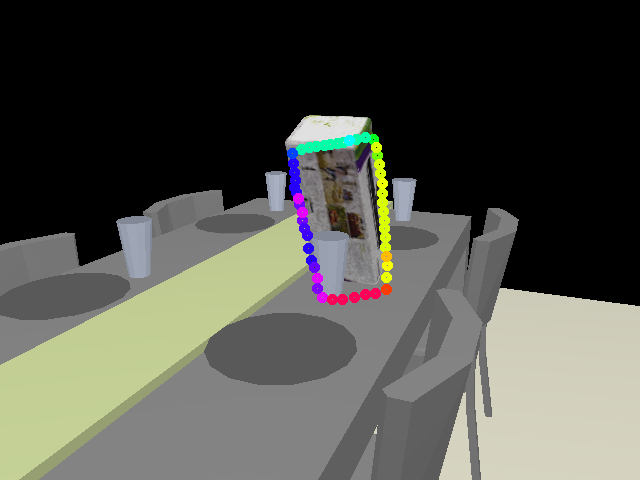} 	
	\caption{Top: Mean translational error for a changing $\lambda$ on every 20th frame for 'kinect\_box' (left) and 'tide' (right). Bottom: Tracking performance on'kinect\_box'. With $\lambda=10^5$, the balance between contour and interior points drives the pose correctly. With $\lambda=10^9$, the energy is dominated by the plane-to-point ICP term, which leads to drifting for planar objects. With an emphasis on contour alone ($\lambda=10$), we deviate later due to an occluding cup. }
	\label{fig:lambda}
\end{figure}

To understand the balancing between contour and interior points, we analyze the influence of a changing $\lambda$. It should both compensate for a different number of sampling points and numerical scale. We fix the sample points to 50 for both modalities to focus solely on the scale difference from the Jacobians. While the ICP values are metric, ranging around $[-1,1]$, the values from the contour Jacobians are in image coordinates and can therefore be in the thousands. 
We chose two sequences, namely 'kinect\_box' and 'tide', and varied $\lambda$. All four sequences are impossible to track via contour alone ($\lambda=10$) since the similarity between foreground and background is too large. On the other hand, relying on a plane-to-point energy alone ($\lambda=10^9$) leads to planar drifting for the 'kinect\_box'. We therefore found $\lambda=10^5$ to be a good compromise (see Figure \ref*{fig:lambda}).

\subsection{Varying the number of sampling points}
\begin{figure}
	\centering
	\includegraphics[width=4cm, height=2.5cm]{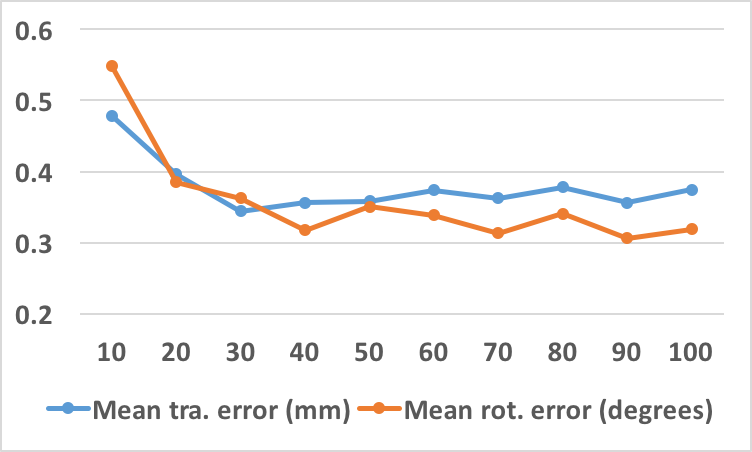} 
	\includegraphics[width=4cm, height=2.5cm]{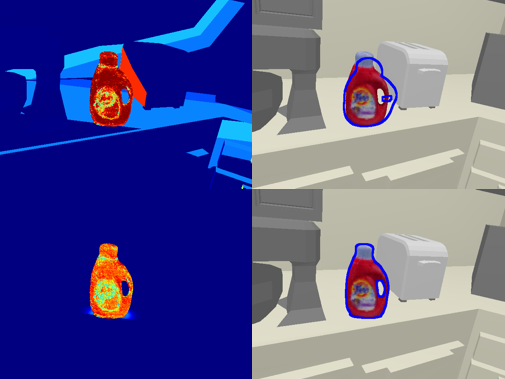} 
	\caption{Left: Average error in translation/rotation for the 'tide' when varying the sample point size. We plot in the same chart since they are similar in scale. Right: Comparison of color posterior vs. cloud-reweighted when tracking with $\lambda=10^3$.}
	\label{fig:points}
\end{figure}

With a fixed $\lambda=10^5$, we now look at the behavior when we change the number of sample points. We chose again the 'tide' since it has rich color and geometry to track against. As can be seen in Figure \ref{fig:points},  we decrease constantly until 30 points where the translational error plateaus while the rotational error decays further, plateauing around 80-90 points. We were surprised to see that a rather small sampling set of 10 contour/interior points already leads to proper energy solutions, enabling successful tracking on the sequence. 

\subsection{Comparison to related work}
We ran our method with $\lambda=10^5$ and $50$ points both on the contour and the interior. Since we wanted to measure the performance of the novel energy alignment with and without the additional cloud weighting, we repeated the experiments for both scenarios. We evaluate accordingly with the others by computing the RMSE on each translational axis as well as each rotational axis. As can be seen from Table \ref{tab:choi}, we outperform the other methods greatly, sometimes even up to one order of magnitude. This result is not really surprising, since we are the only method that does a direct, projective energy minimization. While both C\&C and Krull use a particle filter approach that costs them more than 100ms, Tan evaluates a Random Forest based on depth differences. Tan and C\&C employ depth information only whereas Krull uses RGB-D data like us.

If we compare our runtimes, we are very close to Tan. While they constantly need around 1.5ms, we need less than 3ms on average to compute the full update. If we compute the added cloud weighting, it takes us another 6ms but yields the lowest report error so far on this dataset. Note that both Tan and Krull require a training stage to build their regression structures whereas our method only needs to render 642 views and extract sample information. This takes about 5 seconds in total and requires roughly 10MB per model. Additionally, if we compare to the GPU-enabled dense implementation of Tjaden \etal \cite{Tjaden2016}, we are roughly four times faster on a single CPU core.

\begin{table} 
	\begin{center}
		\begin{tabular}{c|c|c|c|c|c|c|c|}
			\toprule
			\multicolumn{1}{c}{} &
			\multicolumn{1}{c}{} &
			\multicolumn{1}{c}{PCL} &
			\multicolumn{1}{c}{C\&C} &
			\multicolumn{1}{c}{Krull} &
			\multicolumn{1}{c}{Tan} &
			\multicolumn{1}{c}{A} &
			\multicolumn{1}{c}{B} 
			\\
			\midrule
			\multirow{7}{*}{{\rotatebox[origin=c]{90}{(a)~\emph{Kinect Box}}}}
			&$t_x$    &    43.99    &    1.84    & 0.8    &   1.54   &   1.2 & \textbf{0.76} \\
			&$t_y$    &    42.51    &    2.23    & 1.67   &   1.90   &    1.16&  \textbf{1.09} \\
			&$t_z$    &    55.89    &    1.36    & 0.79   &   0.34   &   \textbf{0.30} & 0.38 \\  			
			&\emph{$\alpha$}&    7.62 &    6.41  &  1.11  &    0.42  & \textbf{0.14} & 0.17  \\      		
			&\emph{$\beta$} &    1.87    &    0.76 &        0.55        &  0.22    &    0.23 &  \textbf{0.18}  \\
			&\emph{$\gamma$}&    8.31    &    6.32    &     1.04        &    0.68    &  0.22 & \textbf{0.20} \\
			&\emph{ms}    &    4539    &    166    &     143        &    \textbf{1.5}    &    2.70 &  8.10 \\  			
			\midrule
			\multirow{7}{*}{{\rotatebox[origin=c]{90}{(b)~\emph{Milk}}}}
&$t_x$    &    13.38    &    0.93    & \textbf{    0.51    }    &        1.23        &  0.91  & 0.64 \\
&$t_y$    &    31.45    &    1.94    & 1.27        &        0.74     &    0.71&     \textbf{0.59}\\
&$t_z$    &    26.09    &    1.09    & 0.62        &    \textbf{    0.24    }    &      0.26 & \textbf{0.24    }\\
&\emph{$\alpha$}    &    59.37    &    3.83 &        2.19        &    0.50  &  0.44   &  \textbf{0.41} \\ 
&\emph{$\beta$}    &    19.58    &    1.41 &        1.44        &    \textbf{0.28    } &   0.31 & 0.29 \\
			&\emph{$\gamma$}&    75.03    &    3.26    &     1.90        &   0.46      &  0.43 &  \textbf{0.42} \\
			&    \emph{ms}    &    2205    &    134    &     135        &    \textbf{1.5    }    &    2.72 & 8.54 \\
			
			\midrule
			\multirow{7}{*}{{\rotatebox[origin=c]{90}{(c)~\emph{Orange Juice}}}}
			&    $t_x$    &    2.53    &    0.96    &0.52        &        1.10        &    0.59&  \textbf{0.50} \\ 
			&    $t_y$    &    2.20    &    1.44    & 0.74      &        0.94        &    \textbf{0.64 }  & 0.69 \\
			&    $t_z$    &    1.91    &    1.17    & 0.63        &    0.18   &    0.18 & \textbf{0.17}   \\
			&\emph{$\alpha$}    &    85.81    &    1.32 &        1.28        &       0.35     &  \textbf{ 0.12}  &\textbf{ 0.12}   \\
			&\emph{$\beta$}    &    42.12    &    0.75 &        1.08        &   0.24  & 0.22   &   \textbf{0.20} \\
			&\emph{$\gamma$}&    46.37    &    1.39    &     1.20        &       0.37    & \textbf{0.18} &  0.19 \\
			&    \emph{ms}    &    1637    &    117    &     129        &    \textbf{1.5    }    &     2.79 & 8.79 \\
			
			\midrule
			\multirow{7}{*}{{\rotatebox[origin=c]{90}{(d)~\emph{Tide}}}}
			&    $t_x$    &    1.46    &    0.83    &     0.69      &        0.73        &  0.36 & \textbf{0.34 } \\
			&    $t_y$    &    2.25    &    1.37    & 0.81        &    0.56        &     0.51  &  \textbf{0.49 } \\ 
			&    $t_z$    &    0.92    &    1.20    & 0.81        &    0.24    &       \textbf{0.18} &  \textbf{0.18}\\
			&\emph{$\alpha$}    &    5.15    &    1.78    &     2.10        &    0.31     &  0.20 &  \textbf{0.15 } \\
			&\emph{$\beta$}    &    2.13    &    1.09 &        1.38        &    \textbf{0.25    } &     0.43 & 0.39 \\
			&\emph{$\gamma$}&    2.98    &    1.13    &     1.27        &    \textbf{0.34    }    &  0.39 & 0.37 \\
			&    \emph{ms}    &    2762    &    111    &     116        &    \textbf{1.5    }    &    2.71& 9.42 \\
			
			\midrule
			\multirow{3}{*}{{\rotatebox[origin=c]{90}{\textbf{Mean}}}}
			&    Tra    &    18.72    &    1.36    &     0.82        &        0.81        & 0.58 &\textbf{0.51} \\
			&    Rot    &    29.70    &    2.45    & 1.38        &      0.37        &    0.28  &  \textbf{0.26}\\
			&    ms    &    2786    &    132    & 131        &    \textbf{   1.5    }    &    2.73 & 8.71 \\ 
			
			\bottomrule
		\end{tabular}
		\caption{
			Errors in translation (mm) and rotation (degrees), and the runtime (ms) of the tracking results on the Choi dataset. We compare PCL's ICP, Choi and Christensen (C\&C)~\cite{Choi2013}, Krull~\etal~\cite{Krull2014} and Tan~\etal~\cite{Tan2015} to us without (A) and with cloud weighting (B).
		}
		\vspace{-20pt}
		\label{tab:choi}
	\end{center}
\end{table}

\subsection{Convergence properties}
Since our proposed joint energy has not been applied in this manner before, we were curious about the general convergence behavior. To this end, we used the real-life LineMOD dataset \cite{Hinterstoisser2012}. Although designed for object detection, it has ground truth annotations for 15 textureless objects and we thus mimic a tracking scenario by perturbing the ground truth pose and 'tracking back' to the correct pose. More precisely, we create 1000 perturbations per frame by randomly sampling a perturbation angle in the range [$-\theta$,$\theta$] separately for each axis and a random translational offset in the range [$-t$,$t$] where $t$ is $\frac{1}{10}th$ of the model's diameter. This yields \textbf{more than 1 million runs per sequence and configuration}, giving us a rigorous quantitative convergence analysis which we are presenting in Figure \ref{fig:eval_conv} on 3 sequences\footnote{In the supplement, we present the figures for all sequences.} as histograms over the final rotational error. 
We also plot the mean LineMOD score for each $\theta$. For this, the model cloud is transformed once with the ground truth and with the retrieved pose and if the average Euclidean error between the two is smaller than $\frac{1}{10}th$ of the diameter, we count it as positive. Our optimization is iterative and coarse-to-fine on three levels and we thus computed above score for a different set of iterations. For example 2-2-1 indicates 2 iterations at the coarsest scale, 2 at the middle and 1 at the finest. 

\begin{figure}
	\includegraphics[width=4cm]{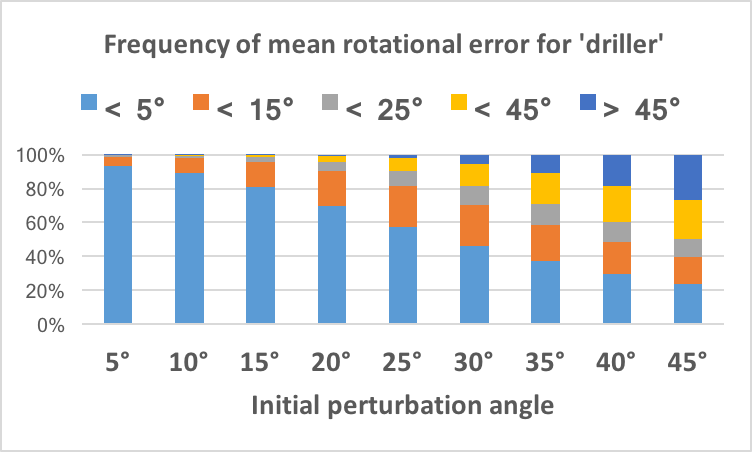} 
	\includegraphics[width=4cm]{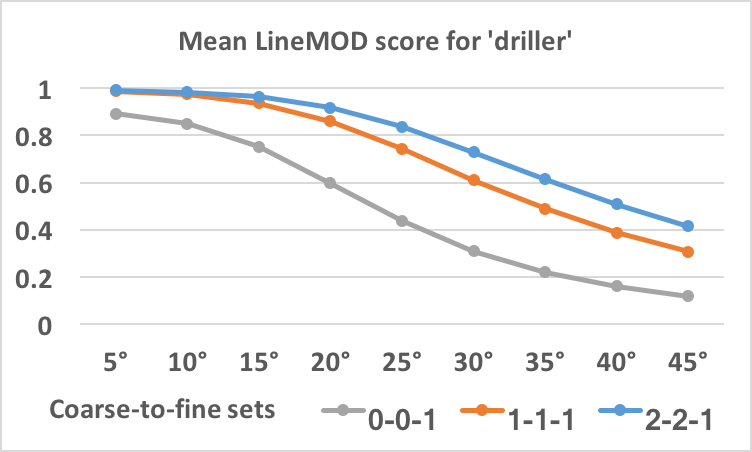}
	\\
	\includegraphics[width=4cm]{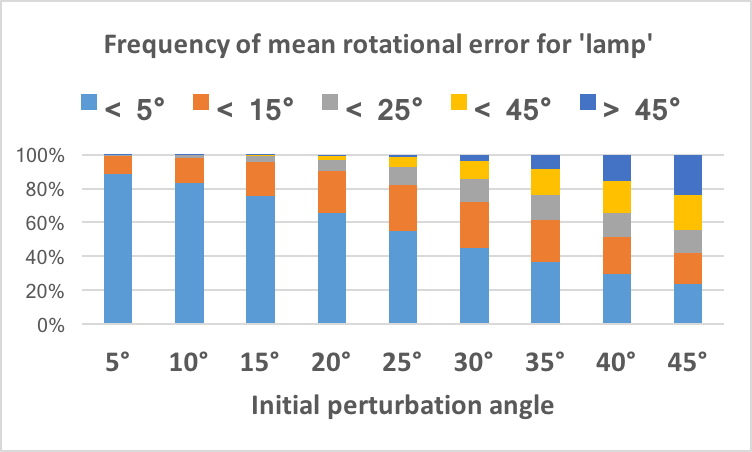}  
	\includegraphics[width=4cm]{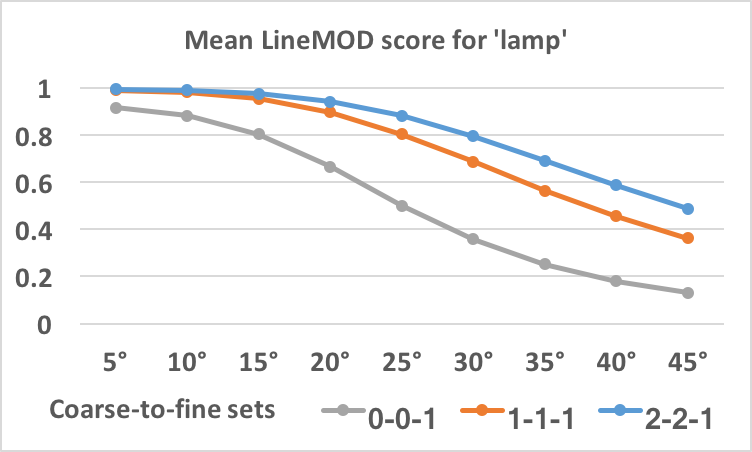}
	\\
	\includegraphics[width=4cm]{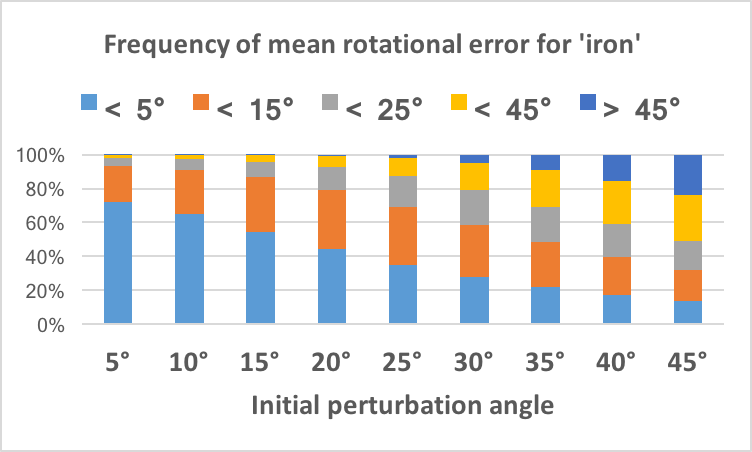}  
	\includegraphics[width=4cm]{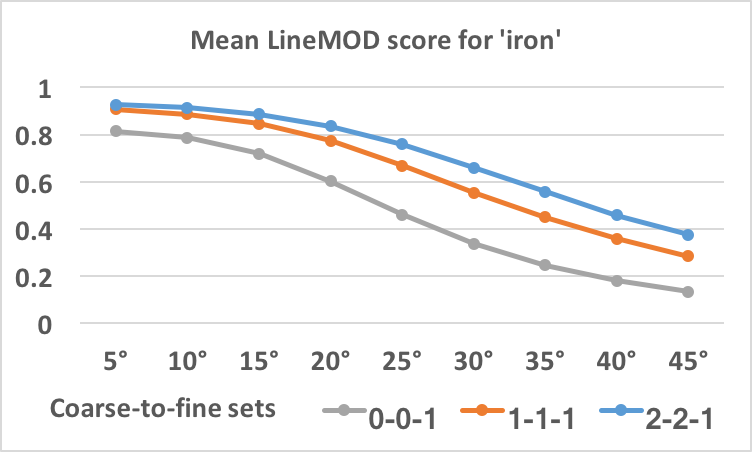}
	\\
	\includegraphics[width=2.7cm]{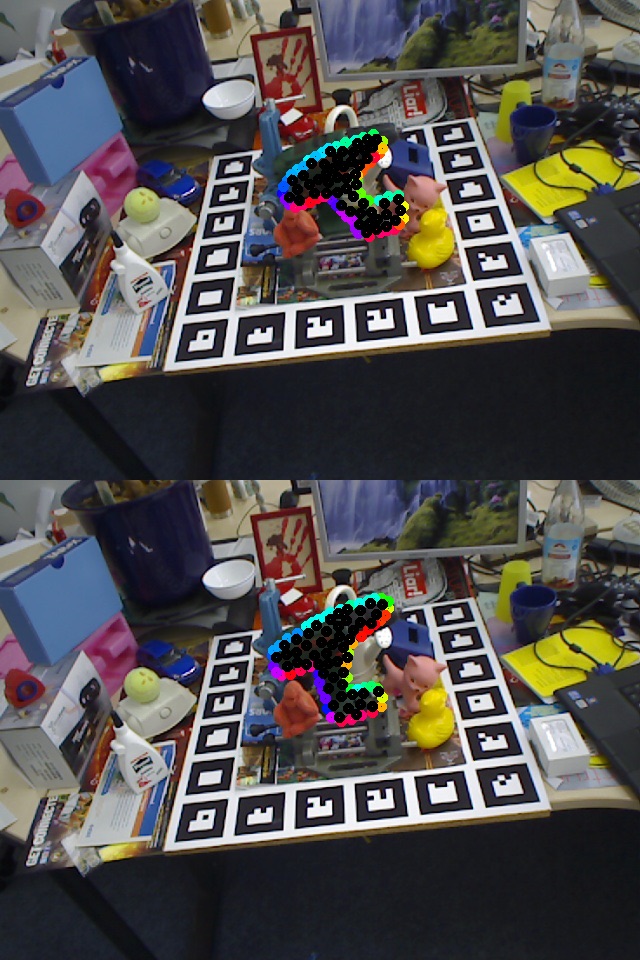} 
	\includegraphics[width=2.7cm]{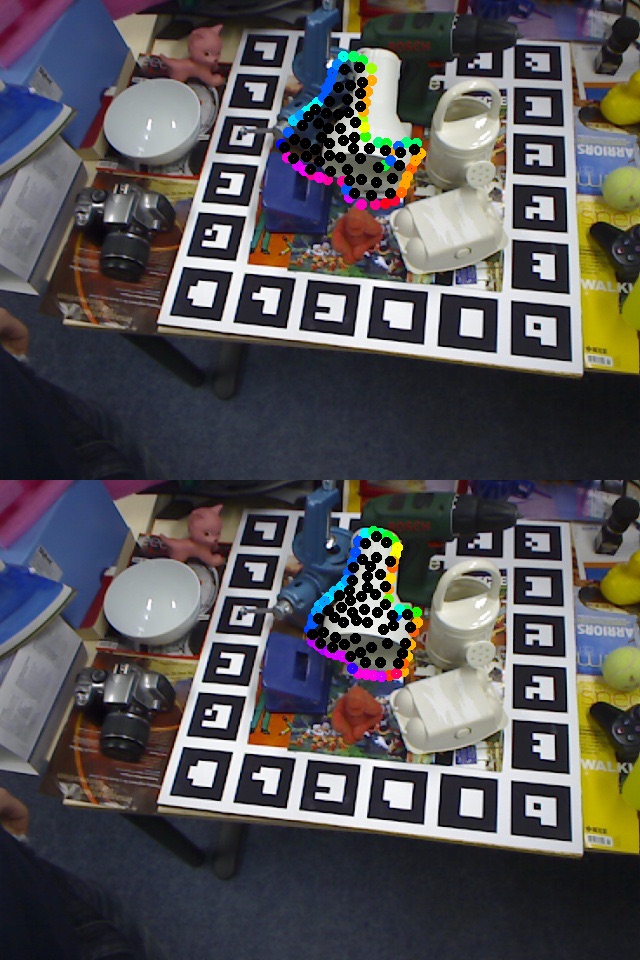}
	\includegraphics[width=2.7cm]{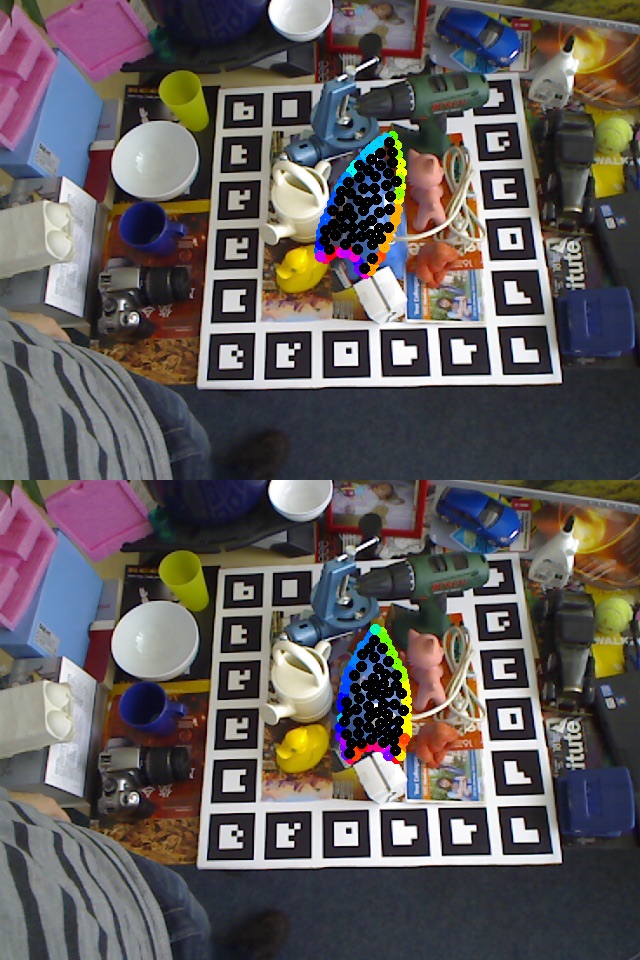}  
	\caption{Top: Relative frequency of rotational error for each $\theta$. Center: Mean LineMOD scores for each $\theta$ and a given iteration scheme. Bottom: Perturbation examples and retrieved poses.}
	\label{fig:eval_conv}
\end{figure}
During tracking a typical change in pose rarely exceeds $5^{\circ}$ on each axis and for this scenario, we can report near-perfect results. Nonetheless, we fare surprisingly well for more difficult pose deviations and degrade gracefully. From the LineMOD scores we see that one iteration on the finest level is not enough to recover stronger perturbations. For very high $\theta$, the additional iterations on the coarser scales can make a difference in up to 10\% which is mainly explained by the SDF rays, capturing larger spatial distances.  

\subsection{Real-data comparison to state-of-the-art}
\begin{figure}
\includegraphics[width=2cm]{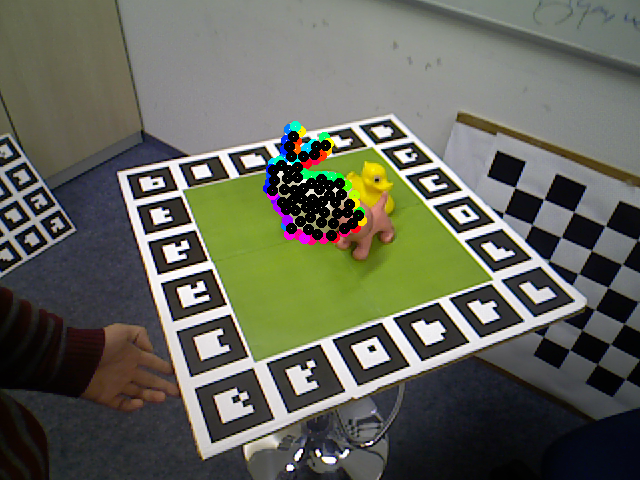} 
\includegraphics[width=2cm]{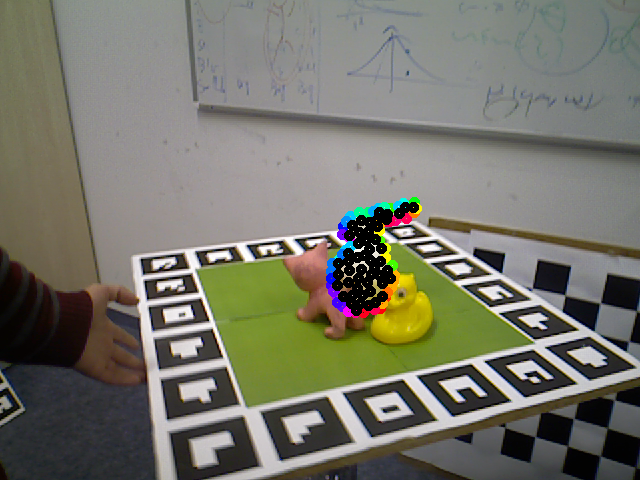} 
\includegraphics[width=2cm]{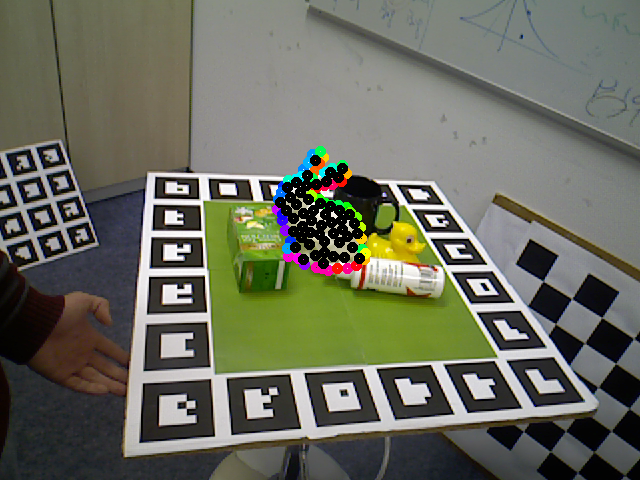} 
\includegraphics[width=2cm]{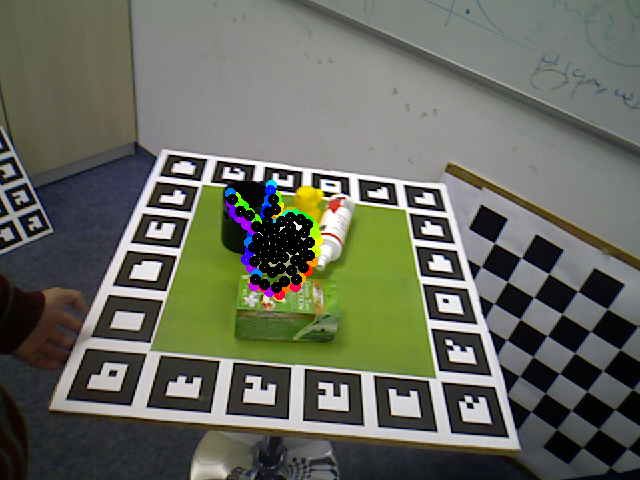} 
\\
\includegraphics[width=4cm]{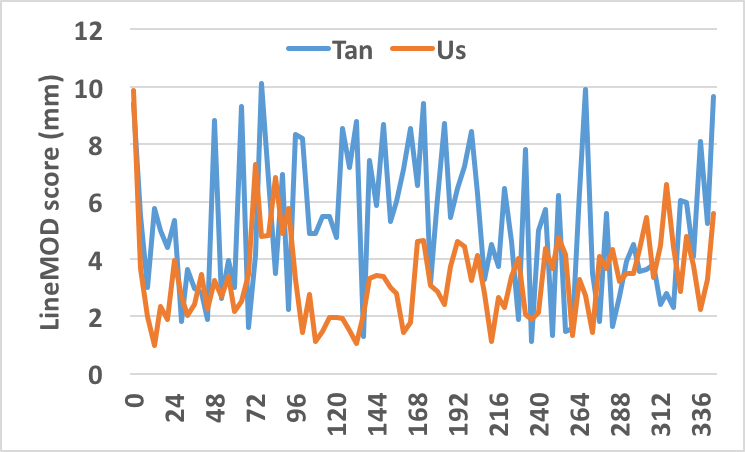} 
\includegraphics[width=4cm]{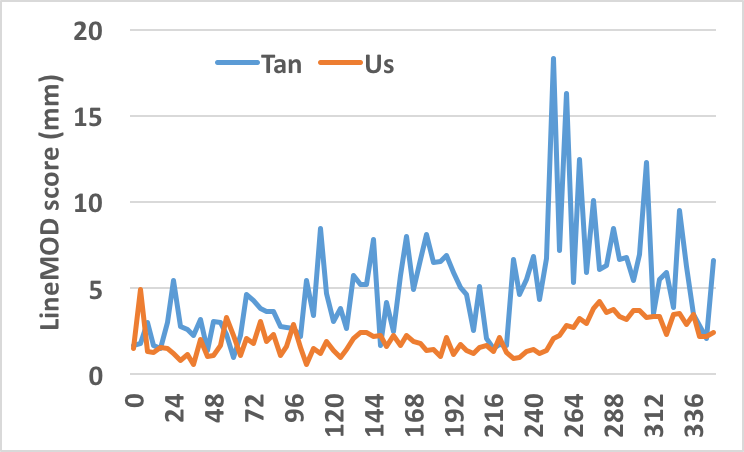} 
\caption{Top: Two frames each from the two sequences that we compared against Tan \etal. Bottom: The LineMOD error for every $4th$ frame on both sequences. We clearly perform better.}
\label{fig:eval_david}
\end{figure}

We thank the authors from Tan \etal for providing two sequences together with ground truth annotation such that we could evaluate our algorithm in direct comparison to their method. In contrast to us, their method has a learned occlusion handling built-in. Both sequences feature a rotating table with a center object to track, undergoing many levels of occlusion. As can be seen from Figure \ref{fig:eval_david} we outperform their approach, especially on the second sequence.

\subsection{Failure cases}
The weakest link in the method is the posterior computation since the whole contour energy is dependent on it. In the case of blur or a sudden change of colors (e.g. illumination) the posterior is misled. Furthermore, with our approximative SDF we sometimes fail for small or non-convex contours where the inner rays are overshooting the interior.

\section{Conclusion}

We have demonstrated how RGB and depth can be utilized in a joint fashion for the goal of accurate and efficient 6D pose tracking. The proposed algorithm relies on a novel optimization scheme that is general enough to be individually applied on either the depth or the RGB modality, while being able to fuse them in a principled way when both are available. Our system runs in real-time using a single CPU core, and can track around 10 objects at 30Hz, which is a realistic upper bound on what can visually fit into one VGA image. At the same time, it is able to report state-of-the-art accuracy and inherent robustness towards occlusion. 

\paragraph{Acknowledgments} The authors would like to thank Henning Tjaden for useful implementation remarks and Toyota Motor Corporation for supporting and funding this work.

{\small
\bibliographystyle{ieee}
\bibliography{egbib}
}

\onecolumn

\begin{center}
{\Huge Supplementary Material}
\end{center}

\section{Convergence on the LineMOD dataset}
We run the experiments on all non-symmetric objects of the LineMOD dataset since the introduced LineMOD measure for symmetric objects is very misleading and allows obvious wrong poses to be regarded as correct. Since we are interested in very accurate poses, these results would not have provided additional insight. As in the main paper, the degradation for larger deviations is increasing but we do not see a sharp sudden decline. Again, additional iterations on multiple levels drastically improves the general alignment under all perturbations.

\begin{figure}[h]
	\includegraphics[width=4.25cm]{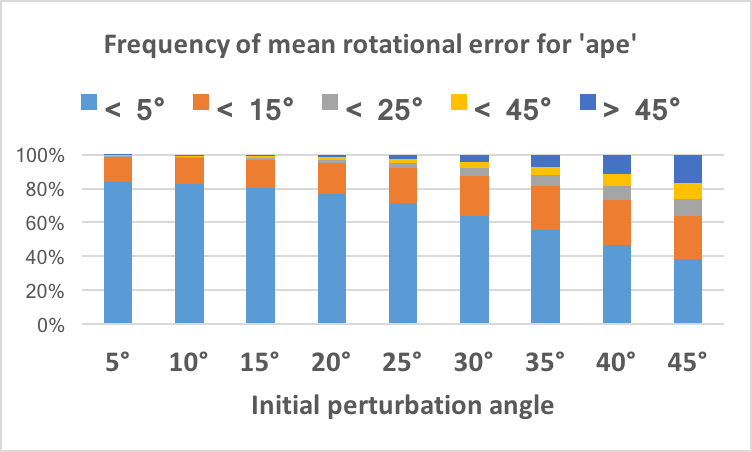} 
	\includegraphics[width=4.25cm]{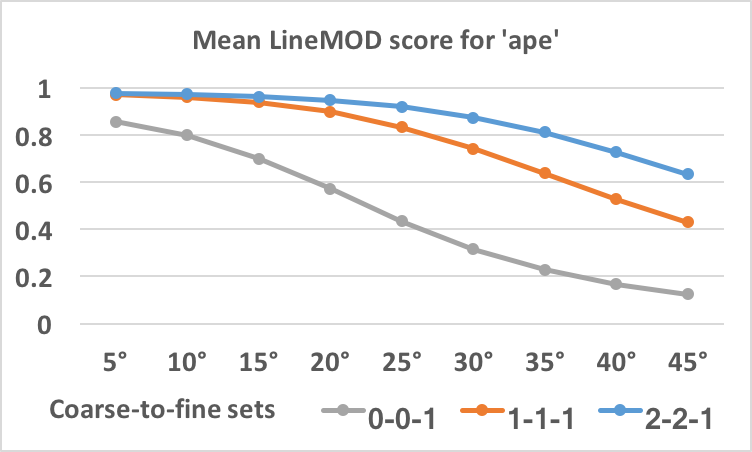}
		\includegraphics[width=4.25cm]{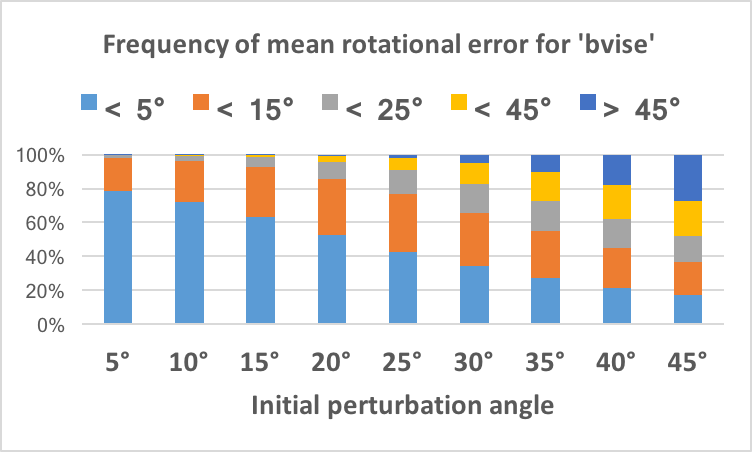} 
		\includegraphics[width=4.25cm]{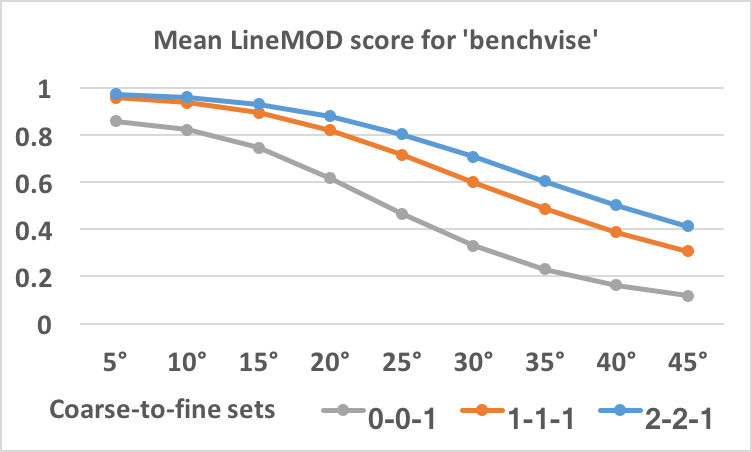}
		\\
					\includegraphics[width=4.25cm]{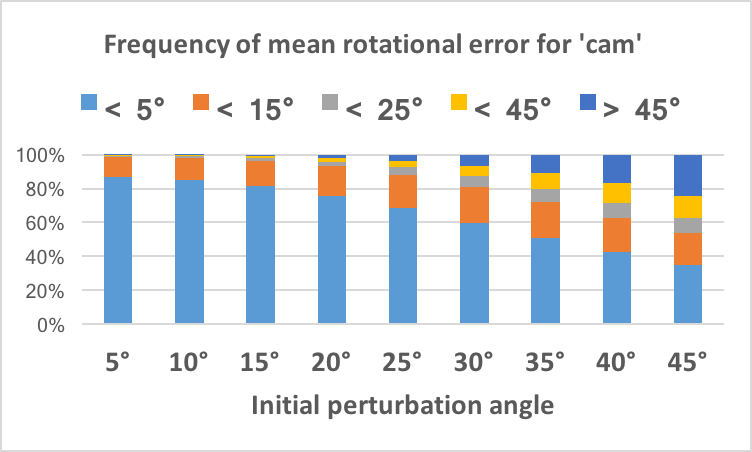} 
					\includegraphics[width=4.25cm]{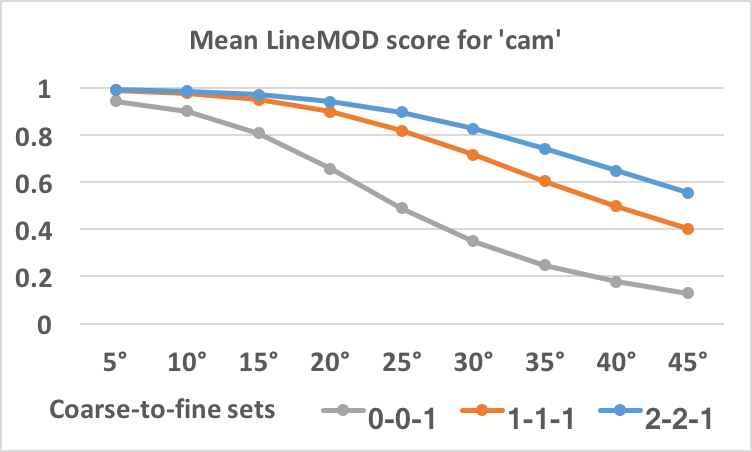}
			\includegraphics[width=4.25cm]{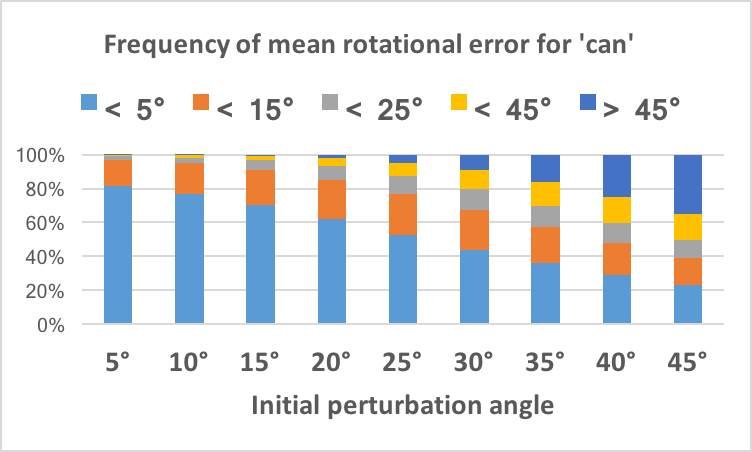} 
			\includegraphics[width=4.25cm]{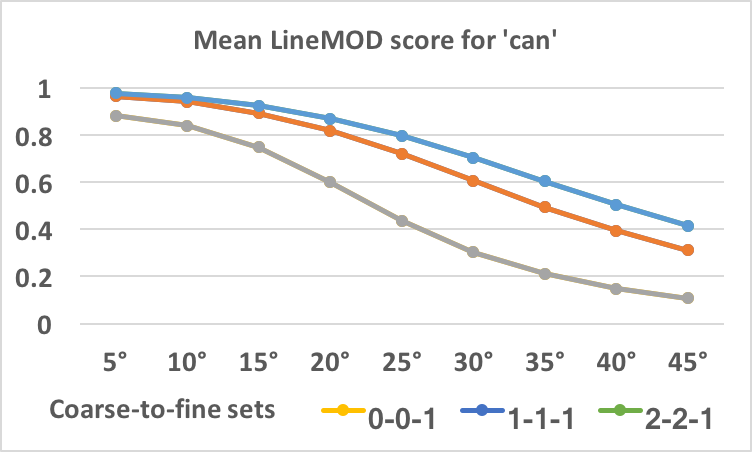}
		\\
		\includegraphics[width=4.25cm]{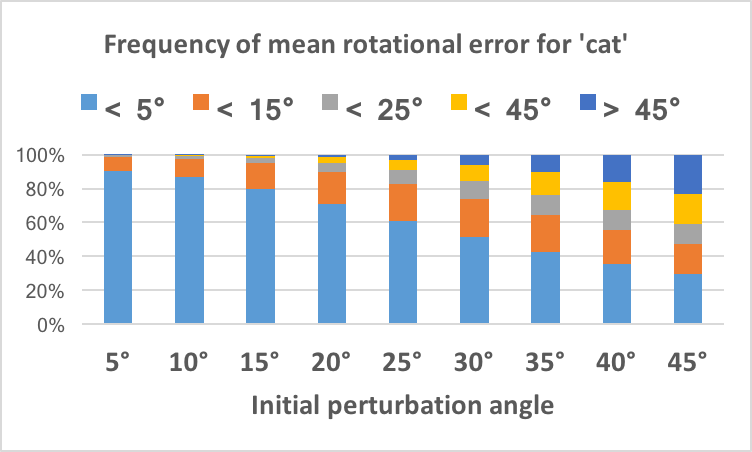} 
		\includegraphics[width=4.25cm]{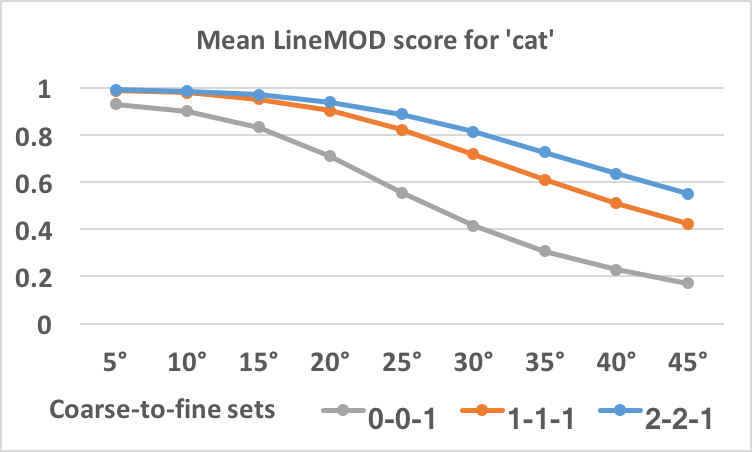}
		\includegraphics[width=4.25cm]{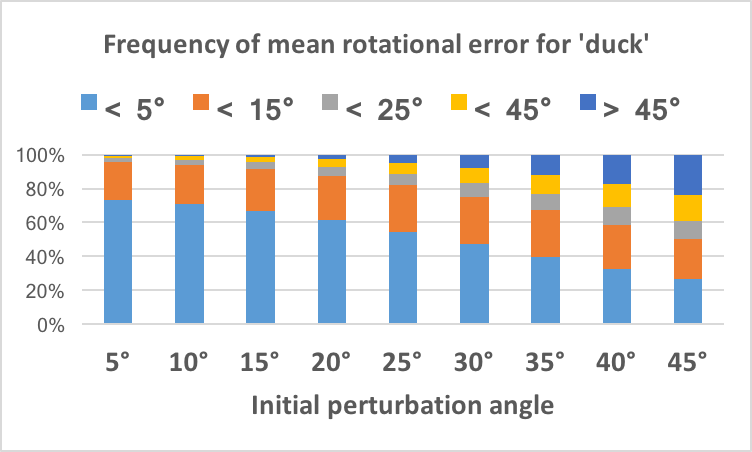} 
		\includegraphics[width=4.25cm]{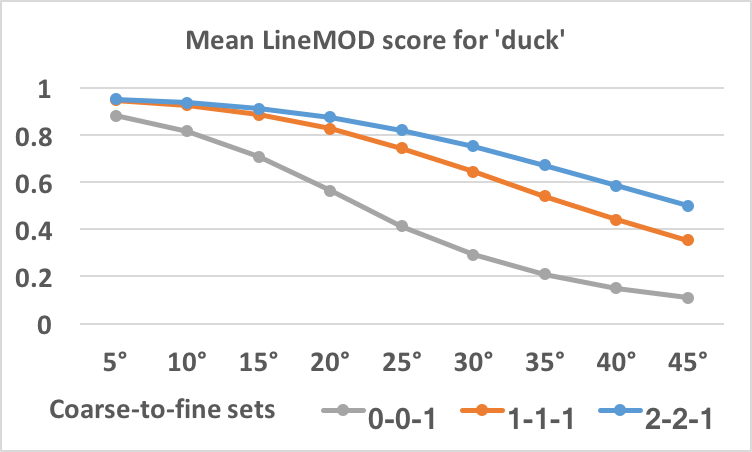}
			\\
\includegraphics[width=4.25cm]{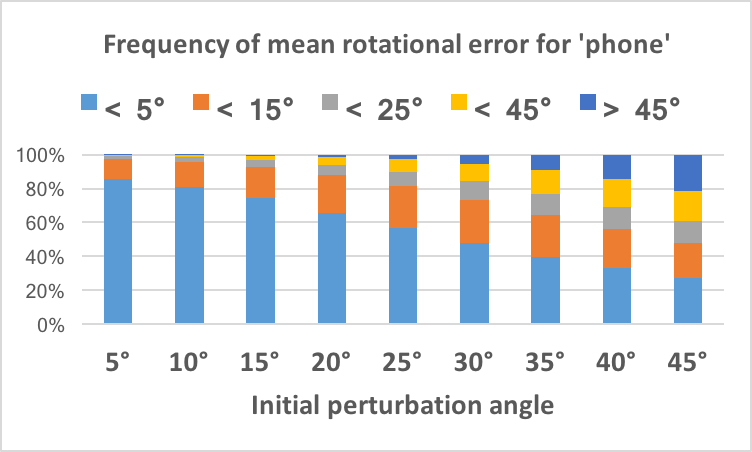} 
\includegraphics[width=4.25cm]{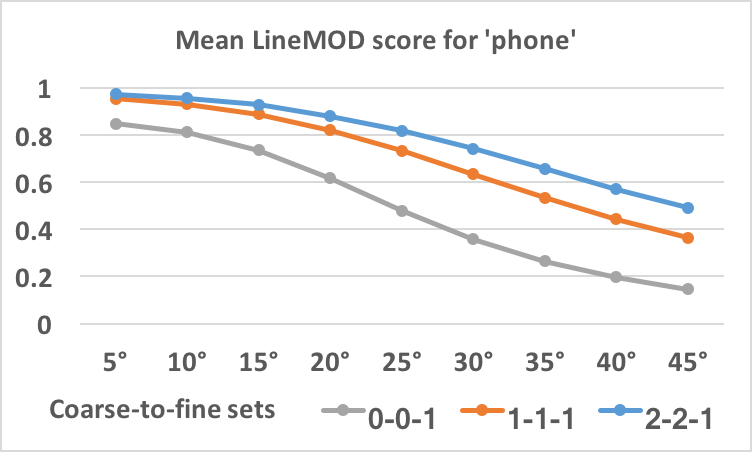}
\includegraphics[width=4.25cm]{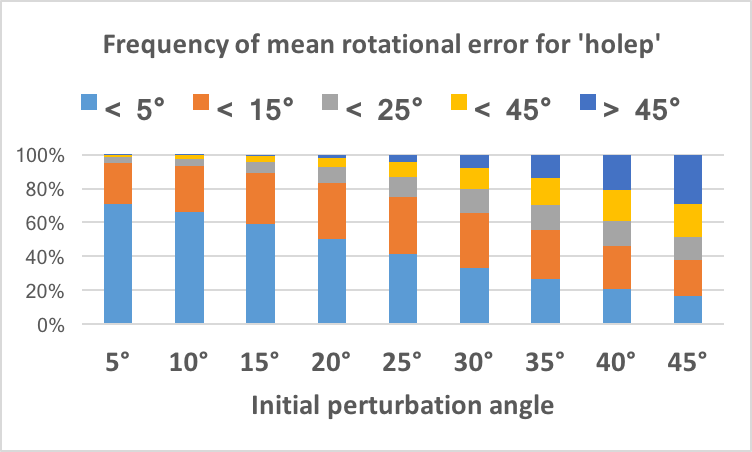} 
\includegraphics[width=4.25cm]{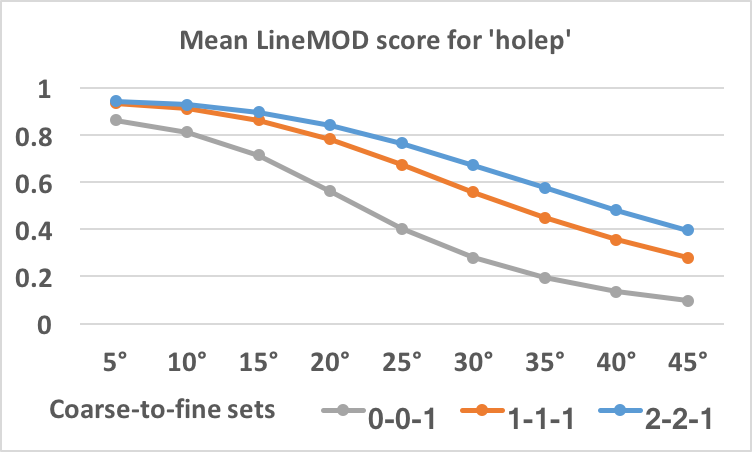}
\end{figure}

\newpage 

\section{Optimization of the Contour Energy}
Starting from the probability of a contour $\phi$ when given an observed image $I$
\begin{equation}
P(\phi|I) := \prod_{x \in \Omega} \bigg(  H_\phi(x) P_f(I(x)) +  (1-H_\phi(x) P_b(I(x)) ) \bigg)
\end{equation}
that we seek to maximize, one can instead minimize the negative log which breaks down to a sum of pixel-wise terms: 
\begin{equation}
 E_{C} := - \sum_{x \in \Omega} \log \bigg(  H_\phi(x)  P_f(I(x)) +  (1-H_\phi(x)  P_b(I(x)) ) \bigg).
\end{equation}
The derivation in respect to a change in pose 
\begin{equation}
\frac{ \partial E_C }{\partial \xi   } = -
\frac{  (P_f - P_b) }{H_\phi (P_f - P_b) + P_b   } 
\frac{ \partial H_\phi }{\partial \phi   }   
\frac{ \partial \phi }{\partial x   }  
\frac{ \partial \pi(X) }{\partial X   } 
\frac{ \partial \Xi(X) }{\partial \xi   } .
\end{equation}
can be then expressed via the following terms. Firstly, the smoothed Heaviside $H_\phi$ and its derivative $\frac{\partial H_\phi}{\partial \phi}$ as a smoothed Dirac delta
\begin{equation}
H_\phi(x) := \frac{1}{\pi} \bigg(-atan(b \cdot x) +  \frac{\pi}{2} \bigg) \hspace{2cm} \frac{\partial H_\phi}{\partial \phi}(x) := \frac{1}{ \pi(x^2 + b^2 + 1)}
\end{equation}
with $b=0.5$ being the grade of applied smoothing to the contour embedding in our implementation. Provided the 3D point $X =(X_x,X_y,X_z)$ to the projected 2D point $x$ and intrinsics $f_x,f_y$, we write the rest of the derivatives as
\begin{equation}
\frac{ \partial \pi(X) }{\partial X   }  = 
 \begin{bmatrix}
 \frac{f_x}{X_z} & 0 & \frac{f_x \cdot X_x} {X_z^2} \\
 0 & \frac{f_y}{X_z} & \frac{f_y \cdot X_y} {X_z^2}  \\
 \end{bmatrix}
  \hspace{2cm}
\frac{ \partial \Xi(X) }{\partial \xi   }  = 
\begin{bmatrix}
1 & 0 & 0 & 0 & X_z & -X_y  \\
0 & 1 & 0 & -X_z & 0 & X_x   \\
0 & 0 & 1 & X_y & -X_x & 0   
\end{bmatrix}.
\end{equation}
Given with the other explanations from the paper, we can now compute the Jacobian $J_x := \frac{ \partial E_C }{\partial \xi   }(x)$  and update the pose as explained via Cholesky decomposition and the exponential map.

\section{Optimization of the Plane-to-Point Energy}
Given source points $s_i$ and source normals $n_i$, we seek an alignment to destination points $d_i$ such that
\begin{equation}
	E_{ICP} := \argmin_{\Xi} \sum_i \bigg ( (\Xi(s_i)  - d_i) \cdot  \Xi_{SO}(n_i)  \bigg ) ^2.
\end{equation}
Deriving in respect to $\xi$ for a given correspondence $i$ yields us
\begin{equation}
\frac{\partial E_{ICP} } { \partial \xi}(s_i, n_i, d_i, \xi) =  2 \cdot ~ \bigg( 
(\Xi(s_i)  - d_i) \cdot \frac{\partial \Xi_{SO} } { \partial \xi}(n_i) ~ + ~ 
\frac{\partial  (\Xi(s_i)  - d_i) } { \partial \xi} \cdot \Xi_{SO}(n_i) \bigg) = 
\end{equation}
\begin{equation}
2 \cdot \bigg( (s_i - d_i) \cdot
\begin{bmatrix}
0 & 0 & 0 & 0 & n_z & -n_y  \\
0 & 0 & 0 & -n_z & 0 & n_x   \\
0 & 0 & 0 & n_y & -n_x & 0   
\end{bmatrix} 
+
\begin{bmatrix}
1 & 0 & 0 & 0 & s_z & -s_y  \\
0 & 1 & 0 & -s_z & 0 & s_x   \\
0 & 0 & 1 & s_y & -s_x & 0   
\end{bmatrix} 
\cdot n_i \bigg)
\end{equation}
Starting from here, we want to employ a Gauss-Newton scheme for the optimization. We thus seek an increment $\Delta \xi $ around $\mathbf{0}$ such that we minimize the error, \ie we conduct Taylor expansion around zero s.t.
\begin{equation}
E_{ICP}(\mathbf{0} + \Delta \xi) = E_{ICP}(\mathbf{0}) + \frac{\partial E_{ICP} } { \partial \xi}(\mathbf{0}) \cdot \Delta \xi + \frac{1}{2!}  (\frac{\partial E_{ICP}^2 } { \partial \xi}(\mathbf{0}) \cdot \Delta \xi) \cdot \Delta \xi + h.o.t.
\end{equation}
Following the typical approximation scheme, we disregard the higher order terms. Defining the residual $r_i := (s_i-d_i) \cdot n_i$ and a Jacobian $J_i := [ ~ n_i^T ~ ~ ~ \big( (s_i-d_i) \times n_i + s_i \times n_i \big)^T ]$, we arrive at
\begin{equation}
E_{ICP}(\mathbf{0} + \Delta \xi) \approx r_i^2 + \Delta \xi^T  J_i^T r_i +   \Delta \xi^T  J_i^T  J_i  \Delta \xi.
\end{equation}
To minimize $E_{ICP}(\mathbf{0} + \Delta \xi)$, we can now derive in respect to $\Delta \xi$ and set it to zero to find the best update $\Delta \xi$:
\begin{equation}
\frac{\partial E_{ICP}(\mathbf{0} + \Delta \xi)}{\partial \Delta \xi } ~ = ~ J_i^T r_i + J_i^T J_i \Delta \xi ~ \stackrel{!}{=} ~  0 ~ \Leftrightarrow ~ \Delta \xi ~ = ~ - (J_i^T J_i)^{-1}  J_i^T r_i .
\end{equation}
Finally, to fuse it seamlessly into the contour optimization, we negate $J_i$ (or alternatively $r_i$) and retrieve the final normal system for the joint energy from the paper.

\section{Derivation of the Posterior}
What we essentially want to formulate is the probability of a model pose $M=[R,t]$ and its projected silhouette $\Omega_f$, given color image $I$ and cloud data $\Pi^{-1}$,
\begin{equation}
P(\Omega_f, M| I, \Pi^{-1}).
\end{equation}
This expression is difficult to compute in general since it involves 2D and 3D entities as well as a 6D pose. Instead, we make the first step towards tractability by assuming that each pixel is independent. We thus rephrase $\Omega_f$ as a binary variable $G \in \{FG,BG\}$ that signifies whether a certain pixel is foreground or background. Dealing now with pixel-wise colors $x$ and cloud points $C$, we also assume that a pose and its projected silhouette are independent entities, given $x$ and $C$:
\begin{equation}
P(G=FG, M |x,C) = P(FG|x,C) \cdot P(M|x,C).
\end{equation}
We now go into the derivation of those two terms. Applying Bayes' rule, we get
\begin{equation}
P(FG|x,C)  :=  \frac{P(x,C|FG) \cdot P(FG) }{ P(x,C) } \hspace{1.5cm}
P(M|x,C)   :=  \frac{P(x,C|M) \cdot P(M) }{ P(x,C) }
\end{equation}
and we now assume further that colors and cloud points are independent, given the foreground or pose model. From here, we marginalize over both instances of $G$ as well as the model pose space:
\begin{equation}
P(FG|x,C) := \frac{P(x|FG) \cdot P(C|FG) \cdot P(FG) }{ \Sigma_{G \in \{FG,BG\} } P(x|G) \cdot P(C|G) \cdot P(G)}
\hspace{0.5cm}
P(M|x,C)  :=  \frac{P(x|M) \cdot P(C|M) \cdot P(M)}{ \int_{\bar{M}} P(x|\bar{M}) \cdot P(C|\bar{M}) \cdot P(\bar{M}) ~ d\bar{M}} .
\end{equation}
We compute $P(x|FG)$ and $P(x|BG)$ from color histograms whereas $P(FG) = \frac{|\Omega_f|}{|\Omega|}$ and $P(BG) = \frac{|\Omega_b|}{|\Omega|}$. While the marginalization over foreground and background is straight-forward, it is intractable for the model pose space. As mentioned in the paper, we assume both $P(x|M)$ and $P(M)$ to be uniform. Furthermore, since the integration over all valid $M$ of our cloud term $P(C|M)$ is constant, we reduce ourselves to a proportionate measure.

\end{document}